\documentclass[letterpaper, 10 pt, conference]{ieeeconf}

\IEEEoverridecommandlockouts
\usepackage{graphicx} 

\usepackage{fancyhdr}
\fancypagestyle{withfooter}{
  
  \fancyhead[L]{}
  \fancyhead[R]{}
  \fancyfoot[C]{\footnotesize Presented at the 2025 IEEE ICRA Workshop on Field Robotics}
}

\title{\LARGE \bf Field Report on Ground Penetrating Radar for Localization at the Mars Desert Research Station}

\author{Anja Sheppard$^{1}$ and Katherine A. Skinner$^{1}$
\thanks{*This work was supported in part by NSF Grant \#DGE 2241144, the University of Michigan Robotics Department, and the University of Michigan Space Institute Pathfinder Grant. This work is also supported in part by NASA Pennsylvania Space Grant Consortium (PSGC) Mini-Grants Program and Duquesne University Biomedical Department funds.}
\thanks{$^{1}$Both authors are with the Department of Robotics,
        University of Michigan, 2505 Hayward St, Ann Arbor, MI 48109, USA.
        {\tt\small \{anjashep, kskin\} @umich.edu}}
}

\begin{document}

\maketitle
\thispagestyle{withfooter}
\pagestyle{withfooter}

\begin{abstract}
In this field report, we detail the lessons learned from our field expedition to collect Ground Penetrating Radar (GPR) data in a Mars analog environment for the purpose of validating GPR localization techniques in rugged environments. Planetary rovers are already equipped with GPR for geologic subsurface characterization. GPR has been successfully used to localize vehicles on Earth, but it has not yet been explored as another modality for localization on a planetary rover. Leveraging GPR for localization can aid in efficient and robust rover pose estimation. In order to demonstrate localizing GPR in a Mars analog environment, we collected over 50 individual survey trajectories during a two-week period at the Mars Desert Research Station (MDRS). In this report, we discuss our methodology, lessons learned, and opportunities for future work.
\end{abstract}

\section{Introduction}

Mars rovers must localize over long periods of time in a GPS-denied environment. Wheel encoders are usually a reliable source of robot odometry, but they fail in high-slip environments such as sandy deposits on Mars. Visual odometry can provide accurate pose estimation, but this method requires good lighting and a lot of computation power. Currently, human operators are still in the loop to hand-correct Perseverance and Curiosity localization predictions due to estimation errors from sensors, although strides have been made to automate the localization process \cite{nash2024}. Upcoming missions such as CADRE \cite{croix2024}, a trio of small robots exploring the lunar surface, will require accurate localization with reduced ground operational support. 

\begin{figure}[ht]
    \centering
    \includegraphics[width=\linewidth]{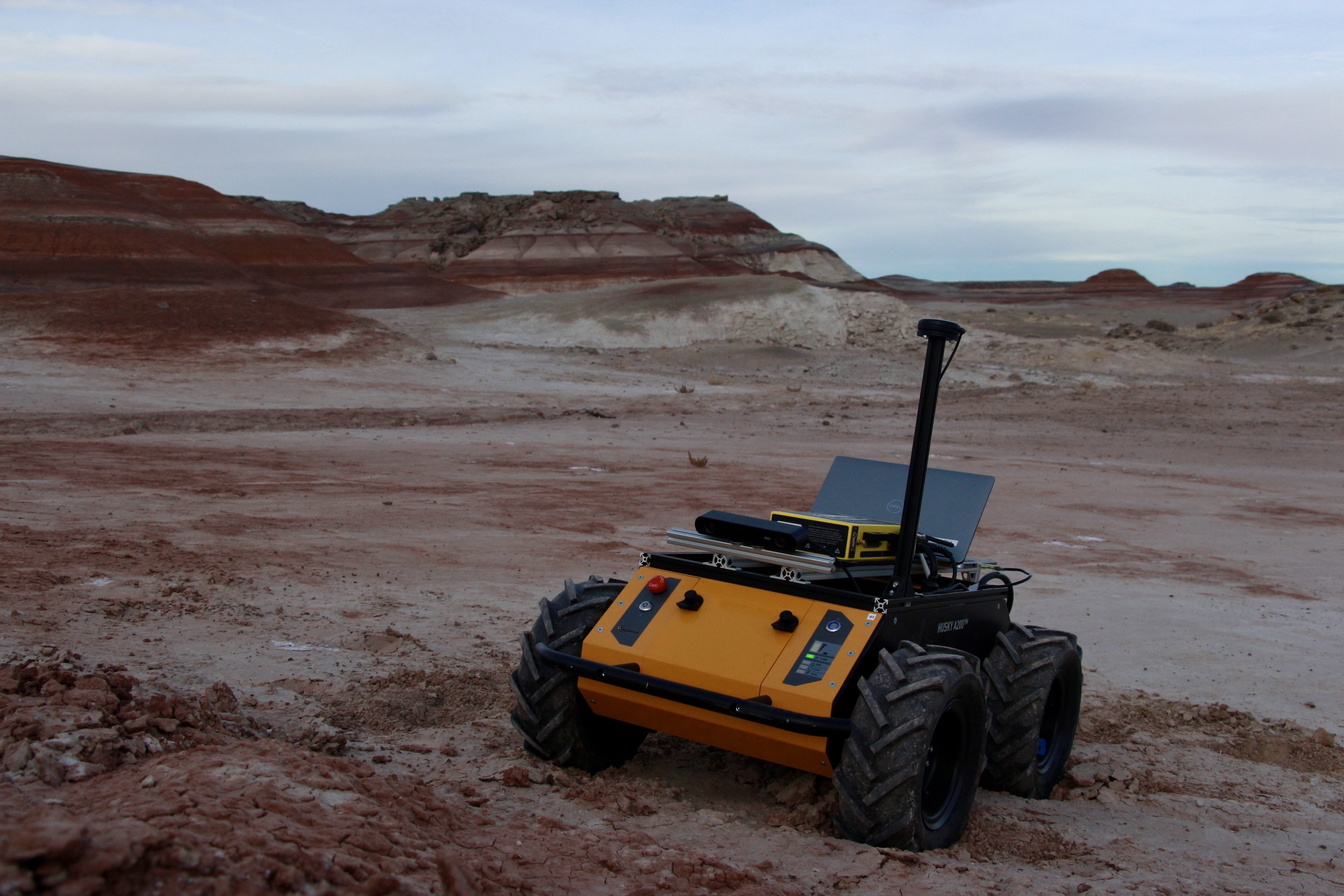}
    \caption{The robot data collection platform at the Mars analog facility: the Mars Desert Research Station. Over the course of two weeks, this platform was used to collect GPR data in a rugged, off-road, Mars analog environment.}
    \label{fig:mdrs}
\end{figure}

On Earth, Ground Penetrating Radar (GPR) has traditionally been used in archaeology, construction, and geology to unobtrusively search for underground features \cite{annan2002}. GPR has also been used to localize autonomous road vehicles either through matching to a pre-registered map of underground features \cite{stanley2013, cornick2016, ort2020, ni2022, ort2023, bi2023, zhang2024, bi2024, xu2024, li2024} or by estimating relative displacement between samples \cite{wickramanayake2022} and then fusing them into a localization framework with other sensors \cite{baikovitz2021}. Underground features are static and do not require good lighting to identify, unlike above-ground images which can suffer from lighting variability.

The Perseverance rover currently on Mars \cite{hamran2020} and future rovers such as ExoMars \cite{ciarletti2017} and CADRE \cite{croix2024} already have GPR sensors on-board. Despite this, there has been no exploration into the usage of this sensor for localization in rugged planetary environments. A major hurdle to exploring this question is the lack of available datasets -- the two existing localizing GPR datasets \cite{ort2021, baikovitz2021cmugpr} are collected on asphalt or concrete surfaces.

In this field report, we present a short overview of our GPR data collection platform (see Fig. \ref{fig:mdrs}) and detail our data collection efforts at a Mars analog environment, the Mars Desert Research Station (MDRS). We discuss the technical details and lessons learned.

\section{Experiments}

\begin{figure}[ht]
    \centering
    \includegraphics[width=\linewidth]{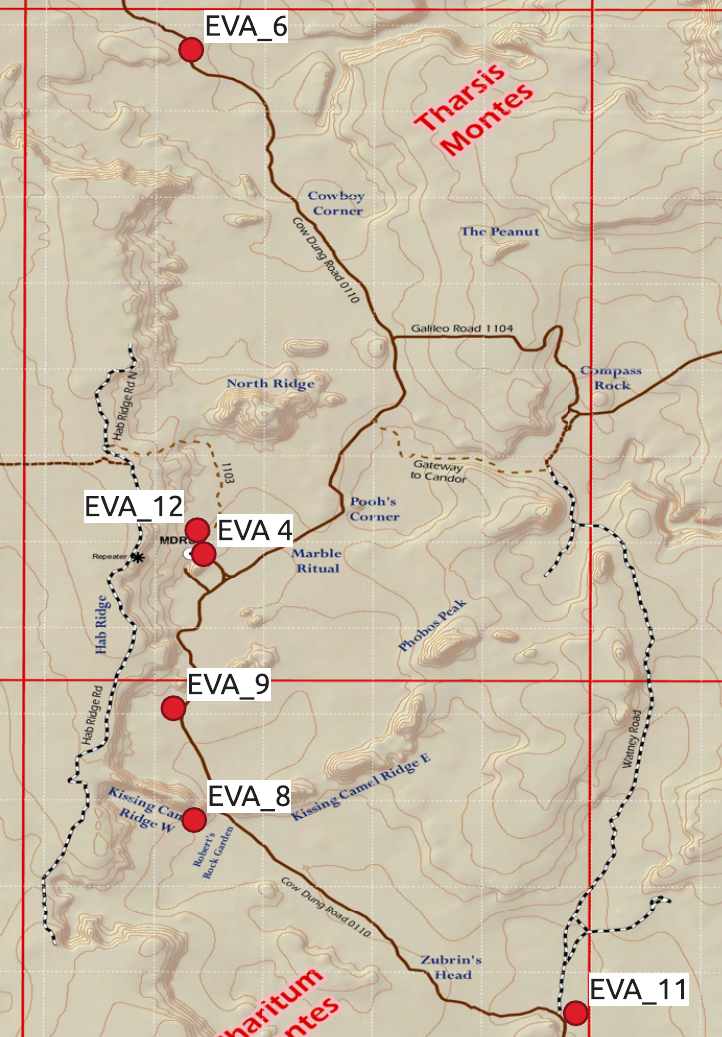}
    \caption{Areas of dataset collection in the region surrounding MDRS in southeast Utah. Base MDRS map provided by the Mars Society.}
    \label{fig:sites}
\end{figure}

The data collection was conducted in January 2024 at MDRS in southeastern Utah. Red soil, lack of vegetation, sandy dunes, and salt presence makes this environment a good geologic analog to Mars \cite{stoker2011}. Over the course of two weeks, six deployments were made and over 50 individual sites were surveyed (see Fig. \ref{fig:sites} for a map of Extravehicular Activity (EVA) locations). 

Our custom robot survey platform is a Clearpath Husky A200, outfitted with a NOGGIN 500 GPR, a Garmin 18x GPS, a ZED2i stereoscopic camera with built-in Inertial Measurement Unit (IMU), and a Jetson Nano for computation. The frequency of a GPR sensor determines the resolution and reach of the sensor, with a higher frequency resulting in higher resolution but less depth penetration. We chose a higher center frequency of 500 MHz as a good tradeoff between resolution and depth.

We deployed a custom Robot Operating System (ROS) driver \cite{sheppard2024} using NOGGIN's SPIDAR SDK for real-time GPR data collection that is synced with the other sensors onboard. The Husky was manually controlled using a Logitech wireless controller to drive in a variety of survey patterns depending on terrain of interest and hazards to the rover.

\begin{figure}[ht]
    \centering
    \includegraphics[width=\linewidth]{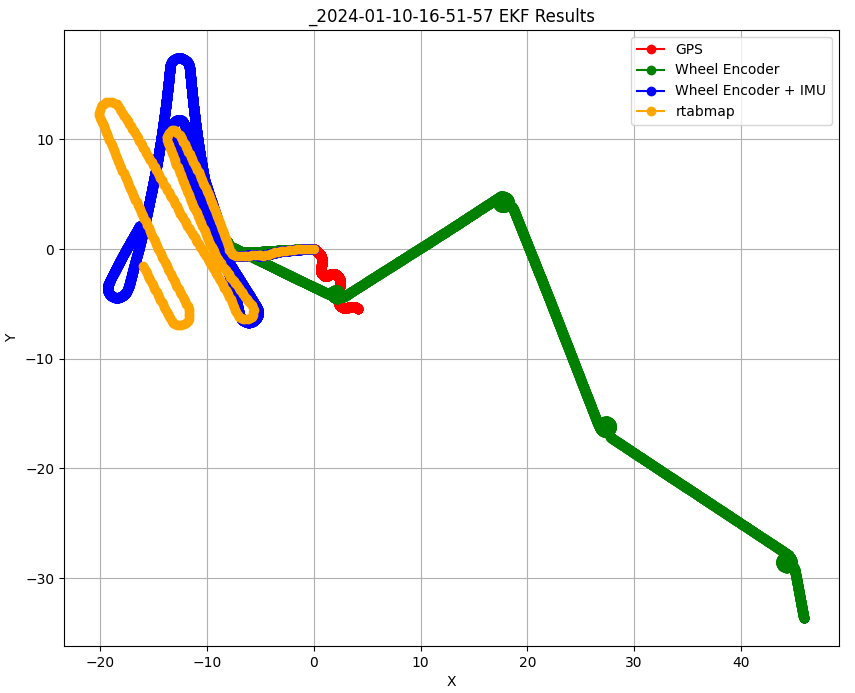}
    \caption{GPS, Extended Kalman Filter (EKF), and RTAB-Map localization results for one survey. Note how the GPS trajectory (in red) is incorrect in scale due to the reported SBAS fixes.}
    \label{fig:sbas}
\end{figure}

\section{Challenges and Lessons Learned}

Deploying robotic survey platforms in extreme environments introduces a host of challenges and requires on-the-spot problem solving. In the winter months in southeastern Utah, the temperatures are low, the sun is bright and low on the horizon, and there is minimal cloud cover. We experienced several days of extremely bright sun, which posed a challenge for the camera on-board the Husky. We found that our camera polarizer film was essential, but even with this addition we often had to drive the robot in stretches perpendicular to the sun azimuth in order to avoid having overly washed out images or images with a strong shadow from the robot.

\begin{figure}[hb]
    \centering
    \includegraphics[width=0.97\linewidth]{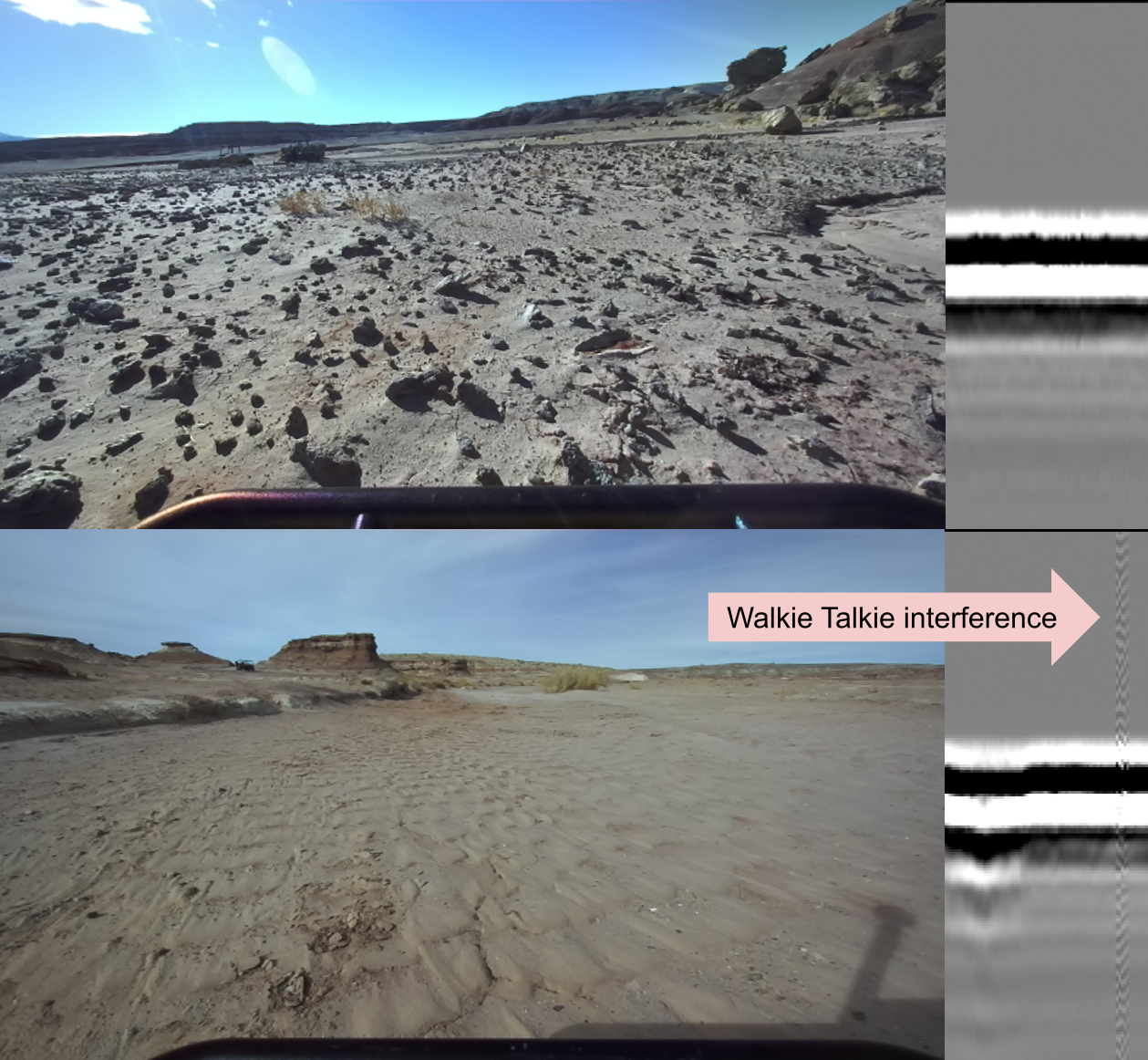}
    \caption{Sample of two sites from the dataset, with camera imagery (left) and GPR radargrams (right). See an example of the walkie talkie interference pointed out by the red arrow.}
    \label{fig:sample}
\end{figure}

During data postprocessing, several additional challenges presented themselves. First, we discovered that an issue with the GPS caused unrecoverable global positioning errors in most of the surveys. Upon inspecting the recorded ROS GPS positions, it seems that when Satellite Based Augmentation System (SBAS) corrections are issued to the GPS position (which typically improves the global position estimate), the reported trajectory is quite noticeably incorrect compared to baselines from the wheel encoder, IMU, or visual odometry, as shown in Fig. \ref{fig:sbas}. We still have not yet identified the source of these errors, as all other reported status data indicated good quality fixes were being obtained. The released dataset contains reference trajectories from RTAB-Map \cite{labbe2013}, a high-quality optimization-based visual odometry localization method that we ran offline, instead of GPS fixes.

Lastly, due to the remoteness of these field expeditions, long-range walkie talkies were used to communicate with a base station for safety and planning purposes. After inspecting the GPR traces, it seems that there was some interference from the walkie talkie transmissions resulting in bursts of more noisy traces (see in Fig. \ref{fig:sample}). In order to mitigate for a variety of noise sources, we had implemented trace stacking, which involves taking several samples in very close succession and then ``stacking'' or averaging them together to get one less noisy sample \cite{utsi2017}. Stacking, along with standard GPR filtering techniques helped to minimize the effects of the walkie talkie interference. We advise that future expeditions pay close attention to the frequency of their long-range communications in order to avoid this.

\section{Conclusion}

In this field report, we present an overview of our field expedition for localizing GPR data collection in a Martian analog environment. Both current and upcoming missions to the Moon and Mars will already have GPR equipped for geological mapping. This work begins to lay the groundwork for leveraging GPR as an additional modality for localization. We detail the logistics of our data collection at a Mars analog environment, and we also present key challenges experienced during our expedition. This dataset has already been used for development of deep learning-based localization methods \cite{sheppard2025}, and future work will include further exploration into multimodal fusion approaches such as a Multimodal Variational Autoencoder \cite{wu2018}.

\section{Acknowledgment}

The authors would like to thank Ben Stanley and the mission support crew at MDRS and the Mars Society for their support during field operations. Additionally, the authors would like to thank Madelyn MacRobbie, Rebecca McCallin, Ben Kazimer, Anna Tretiakova, and Wing Lam Chan for their assistance in data collection in the field.

\bibliographystyle{ieeetr}
\bibliography{main}

\end{document}